\documentclass[11pt,a4paper]{article}
\usepackage[hyperref]{acl2019}
\usepackage{times}
\usepackage{latexsym}
\usepackage{url}

\usepackage{fancyvrb, adjustbox}
\usepackage{bbold}
\usepackage{url}
\usepackage[utf8]{inputenc}
\usepackage{array,multirow}
\aclfinalcopy

\newcommand\blfootnote[1]{%
	\begingroup
	\renewcommand\thefootnote{}\footnote{#1}%
	\addtocounter{footnote}{-1}%
	\endgroup
}

\title{Dialog State Tracking: A Neural Reading Comprehension Approach}
\author{Shuyang Gao*, Abhishek Sethi*, Sanchit Agarwal* \\
 \textbf{Tagyoung Chung}, \textbf{Dilek Hakkani-Tur} \\
Amazon Alexa AI \\
\texttt{\{shuyag, abhsethi, agsanchi, tagyoung, hakkanit\}@amazon.com} \\
 }
\date{}

\begin{document}
\maketitle
\begin{abstract}
Dialog state tracking is used to estimate the current belief state of a dialog given
all the preceding conversation. Machine reading comprehension, on the other hand,
focuses on building systems that read passages of text and answer questions that
require some understanding of passages. We formulate dialog state
tracking as a reading comprehension task to answer the question \textit{what is the state
of the current dialog?} after reading conversational context. In contrast to
traditional state tracking methods where the dialog state is often predicted as
a distribution over a closed set of all the possible slot values within an
ontology, our method uses a simple attention-based neural network to point to
the slot values within the conversation. Experiments on MultiWOZ-2.0
cross-domain dialog dataset show that our simple system can obtain similar
accuracies compared to the previous more complex methods. By exploiting
recent advances in contextual word embeddings, adding a model that
explicitly tracks whether a slot value should be carried over to the next turn,
and combining our method with a traditional joint state tracking method that relies on closed set vocabulary, we can obtain a joint-goal accuracy of 47.33\% on the standard test split, exceeding current state-of-the-art by 11.75\%**\@.
\end{abstract}

\blfootnote{*Authors contributed equally.}
\blfootnote{**We note that after publication, a new state-of-the-art can now be obtained with a similar attention mechanism followed by a enoder-decoder architecture~\cite{wu2019transferable}.}
\section{Introduction}
\label{sec:intro}
%Introduce the dialog state representation here. Check Rastogi's paper: he has a good description. Put an example from the training data here with the respective dialog acts. Then discuss the discriminative approach to DST and contrast it with a generative approach where we learn distribution over the slot values. Use Rahul's paper as reference for this section.    

A task-oriented spoken dialog system involves continuous interaction with a
machine agent and a human who wants to accomplish a predefined task through
speech. Broadly speaking, the system has four components, the Automatic Speech
Recognition (ASR) module, the Natural Language Understanding (NLU) module, the
Natural Language Generation (NLG) module, and the Dialog Manager. The dialog
manager has two primary missions: dialog state tracking (DST) and decision
making. At each dialog turn, the state tracker updates the belief state based on
the information received from the ASR and the NLU modules. Subsequently, the
dialog manager chooses the action based on the dialog state, the dialog policy
and the backend results produced from previously executed actions.

Table~\ref{table:conv} shows an example conversation with the associated dialog
state. Typical dialog state tracking system combines user speech, NLU output,
and context from previous turns to track what has happened in a dialog. More
specifically, the  dialog  state  at  each  turn  is defined  as a  distribution
over a set of predefined variables~\cite{williams2005factored}. The
distributions output by a dialog state tracker are sometimes referred to as the
tracker’s belief or the belief state. Typically, the tracker has complete access
to the history of the dialog up to the current turn.

Traditional machine learning approaches to dialog state tracking have two forms,
generative and discriminative. In generative approaches, a dialog is modeled as
a dynamic Bayesian network where true dialog state and true user action are
unobserved random variables~\cite{williams2007partially}; whereas the
discriminative approaches are directly modeling the distribution over the dialog
state given arbitrary input features. 

Despite the popularity of these approaches, they often suffer from a common
yet overlooked problem --- relying on fixed ontologies. These systems, therefore, have
trouble handling previously unseen mentions. On the other hand, reading
comprehension tasks~\cite{rajpurkar2016squad, chen2017reading, reddy2019coqa}
require us to find the answer spans within the given passage and hence
state-of-the-art models are developed in such a way that a fixed vocabulary for
an answer is usually not required. Motivated by the limitations of previous
dialog state tracking methods and the recent advances in reading
comprehension~\cite{chen2018neural}, we propose a reading comprehension based
approach to dialog state tracking. In our approach, we view the dialog as a
\textit{passage} and ask the question \textit{what is the state of the current
dialog?} We use a simple attention-based neural network model to find answer
spans by directly pointing to the tokens within the dialog, which is similar to
\citet{chen2017reading}. In addition to this attentive reading model, we also
introduce two simple models into our dialog state tracking pipeline, a
\textit{slot carryover} model to help the tracker make a binary decision whether
the slot values from the previous turn should be used; a \textit{slot type} model
to predict  whether the answer is \{\texttt{Yes}, \texttt{No},
\texttt{DontCare}, \texttt{Span}\}, which is similar to \citet{zhu2018sdnet}. To
summarize our contributions:
\begin{itemize}
\item We formulate dialog state tracking as a reading comprehension task and
propose a simple attention-based neural network to find the state answer as a
span over tokens within the dialog. Our approach overcomes the limitations of
fixed-vocabulary issue in previous approaches and can generalize to unseen state
values.
\item We present the task of dialog state tracking as making three sequential
decisions: i) a binary \textit{carryover} decision by a simple slot carryover
model ii) a \textit{slot type} decision by a slot type model iii) a
\textit{slot span} decision by an attentive reading comprehension model. We show
effectiveness of this approach. 
\item We adopt recent progress in large pre-trained contextual word embeddings,
i.e., BERT~\cite{devlin2018bert} into dialog state tracking, and get
considerable improvement. 
\item We show our proposed model outperforms more complex previously published methods on
the recently released MultiWOZ-2.0 corpus~\cite{budzianowski2018multiwoz,
ramadan2018large}. Our approach achieves a joint-goal accuracy of 42.12\%,
resulting in a 6.5\% absolute improvement over previous state-of-the-art.
Furthermore, if we combine our results with the traditional joint state tracking
method in~\citet{liu2017end}, we achieve a joint-goal accuracy of 47.33\%,
further advancing the state-of-the-art by 11.75\%. 
\item We provide an in-depth error analysis of our methods on the MultiWOZ-2.0 dataset
and explain to what extent an attention-based reading comprehension model
can be effective for dialog state tracking and inspire future improvements on this model.

\end{itemize}

\begin{table}[] \small
    \begin{tabular}{ll} \small
		User:  & I need to book a hotel in the east that has 4 stars.                    \\
		\verb+Hotel+     & \verb+area=east, stars=4+                            \\
		Agent: &I can help you with that. What is your price range?				\\

		User:  & That doesn't matter if it has free wifi and parking.            \\
		\verb+Hotel+ & \verb+parking=yes, internet=yes+         \\
		& \verb+price=dontcare, stars=4, area=east+         \\
		Agent: & If you'd like something cheap,             \\
		&  I recommend  Allenbell \\ 
		User:  & That sounds good,  I would also like a  \\
		&taxi to the hotel from cambridge \\
		\verb+Hotel+ & \verb+parking=yes, internet=yes+        \\
		& \verb+price=dontcare, area=east, stars=4+        \\
		\verb+Taxi+      &  \verb+departure=Cambridge+        \\
		&  \verb+destination=Allenbell+        \\ 
	\end{tabular}
	\caption{An example conversation in MultiWOZ-2.0 with dialog states after each turn.}\label{table:conv}
\end{table}

\section{Related Work}
\paragraph{Dialog State Tracking} 
Traditionally, dialog state tracking methods assume a \textit{fixed ontology},
wherein the output space of a slot is constrained by the predefined set of
possible values~\cite{liu2017end}. However, these approaches are not applicable
for unseen values and do not scale for large or potentially unbounded vocabulary
\cite{nouri2018toward}. To address these concerns, a class of methods employing
scoring mechanisms to predict the slot value from a endogenously defined set of
candidates have been proposed \cite{rastogi2017scalable, goel2018flexible}. In
these methods, the candidates are derived from either a predefined ontology or
by extraction of a word or $n$-grams in the prior dialog context. Previously,
\citet{perez2017dialog} also formulated state tracking as a machine reading
comprehension problem. However, their model architecture used a memory network
which is relatively complex and still assumes a fixed-set vocabulary. Perhaps,
the most similar technique to our work is the pointer networks proposed
by~\citet{xu2018end} wherein an attention-based mechanism is employed to
\textit{point} the start and end token of a slot value. However, their
formulation does not incorporate a \textit{slot carryover} component and
outlines an encoder-decoder architecture in which the slot type embeddings are
derived from the last state of the RNN. 

\paragraph{Reading Comprehension} 
% Discuss CoQA, SQuAD, Hotpot QA and their tasks. Cite them. Discuss the underlying principle and show that the same principle can extend to dialog state tracking problem. Discuss BIDEF and Deep QA models to show the current research in the area. Use the PhD thesis and recent papers by Danqi Chen.   
A reading comprehension task is commonly formulated as a supervised learning
problem where for a given training dataset, the goal is to learn a predictor,
which takes a passage $p$ and a corresponding question $q$ as inputs and gives
the answer $a$ as output. In these tasks, an answer type can be cloze-style as
in CNN/Daily Mail \cite{hermann2015teaching}, multiple choice as in MCTest
\cite{richardson2013mctest}, span prediction as in SQuaD
\cite{rajpurkar2016squad}, and free-form answer as in
NarrativeQA~\cite{kovcisky2018narrativeqa}. In span prediction tasks, most
models encode a question into an embedding and generate an embedding for each
token in the passage and then a similarity function employing attention
mechanism between the question and words in the passage to decide the starting
and ending positions of the answer spans \cite{chen2017reading, chen2018neural}.
This approach is fairly generic and can be extended to multiple choice questions
by employing bilinear product for different types~\cite{lai2017race} or to
free-form text by employing seq-to-seq models~\cite{sutskever2014sequence}. 

\paragraph{Deep Contextual Word Embeddings}
The recent advancements in the neural representation of words includes using
character embeddings~\cite{seo2016bidirectional} and more recently using
contextualized embeddings such as ELMO~\cite{peters2018deep} and
BERT~\cite{devlin2018bert}. These methods are usually trained on a very large
corpus using a language model objective and show superior results across a
variety of tasks. Given their wide applicability \cite{liu2019linguistic}, we
employ these architectures in our dialog state tracking task.

\section{Our Approach}
\label{sec:model} 

\subsection{DST as Reading Comprehension}
Let us denote a sub-dialog $D_t$ of a dialog $D$ as prefix of a full dialog ending
with user's $t$th utterance, then state of the dialog $D_t$ is defined by the
values of constituent slots $s_j(t)$, i.e., $S_t = \{s_1(t), s_2(t), .s_j(t),
\dots, s_M(t)\}$. 

Using the terminology in reading comprehension tasks, we can treat $D_t$ as a
\textit{passage}, and for each slot $i$, we formulate a question $q_i$:
\textit{what is the value for slot $i$?} The dialog state tracking task then
becomes understanding a sub-dialog $D_t$ and to answer the question $q_i$ for
each slot $i$.

\begin{figure*}[t!]
	\centering
		\includegraphics[width=1.0\textwidth]{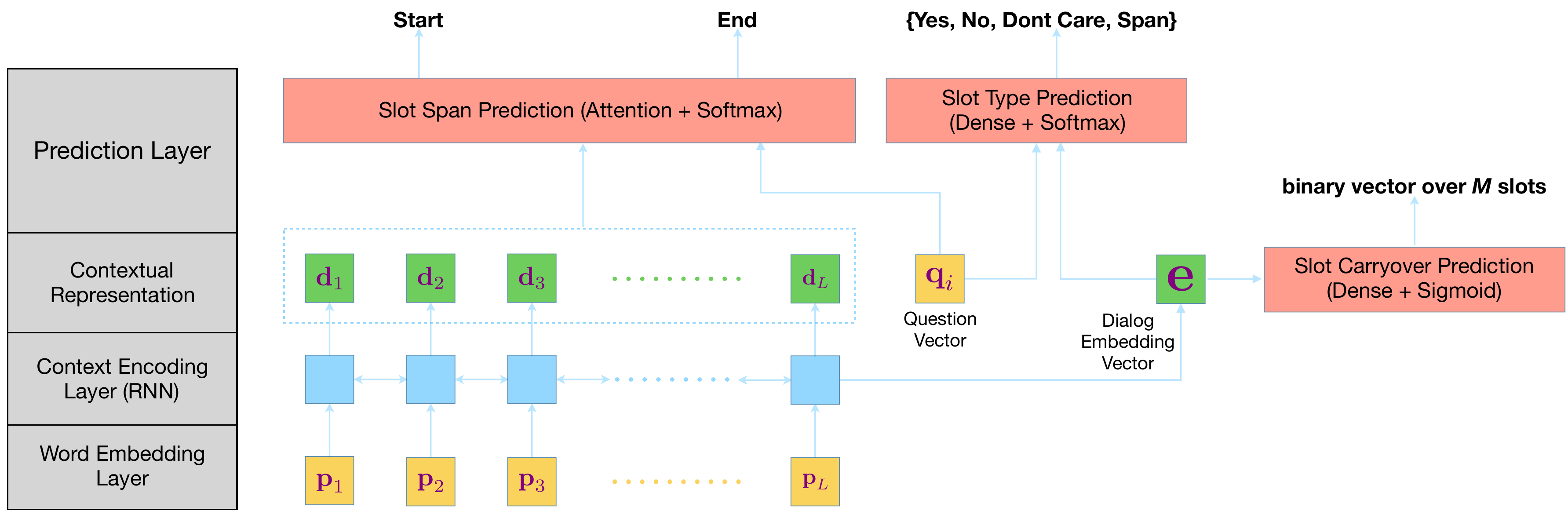} % first figure itself
		\caption{Our attentive reading comprehension system for dialog state
tracking. There are three prediction components on top (from right to left): 1)
slot carryover model to predict whether a particular slot needs to be updated
from previous turn 2) slot type model to predict the type of slot values from
\{\texttt{Yes}, \texttt{No}, \texttt{DontCare}, \texttt{Span}\} 3) slot span model to predict the start and end span of the value within the dialog.}\label{fig:architecture}
\end{figure*}

\subsection{Encoding} 
\paragraph{Dialog Encoding}
For a given dialog $D_t$ at turn $t$, we first concatenate user utterances and
agent utterances $\{\textbf{u}_1, \textbf{a}_1, \textbf{u}_2, \textbf{a}_2,
\dots, \textbf{u}_t\}$. To differentiate between user utterance and agent
utterance, we add symbol \texttt{[U]} before each user utterance and
\texttt{[A]} before each agent utterance. Then, we use pre-trained word vectors
to form $\textbf{p}_i$ for each token in the dialog sequence and pass them as
input into a recurrent neural network, i.e.,
\begin{equation}
 \{\textbf{d}_1, \textbf{d}_2, \dots \textbf{d}_{L} \}= {RNN} (\textbf{p}_1, \textbf{p}_2, \dots {\textbf{p}}_{L})
\end{equation}

\noindent where $L$ is the total length of the concatenated dialog sequence and
$\textbf{d}_i$ is the output of RNN for each token, which is expected to encode
context-aware information of the token. In particular, for pre-trained word
vectors $\textbf{p}_i$, we experiment with using deep contextualized word
embeddings using BERT~\cite{devlin2018bert}. For RNN, we use a one layer
bidirectional long short-term memory network (LSTM) and each $\textbf{d}_i$ is
the concatenation of two LSTMs from both directions, i.e., $\textbf{d}_i =
(\overleftarrow{\textbf{d}_i};\overrightarrow{\textbf{d}_i})$. Furthermore, we
denote $\textbf{e}(t)$ as our dialog embedding at turn $t$ as follows:
\begin{equation}
    \textbf{e}(t) = (\overleftarrow{\textbf{d}_1}; \overrightarrow{\textbf{d}_{L}})	
\end{equation}
 
\paragraph{Question Encoding}
In our methodology, we formulate questions $q_i$ defined earlier as \textit{what
is the value for slot i?} For each dialog, there are $M$ similar questions
corresponding to $M$ slots, therefore, we represent each question $q_i$ as a
fixed-dimension vector $\textbf{q}_i$ to learn. 

\subsection{Models}
\paragraph{Overview}
In our full model set up, three different model components are used to make a sequence
of predictions: first, we use a \textit{slot carryover} model for deciding
whether to carryover a slot value from the last turn. If the first model decided
not to carry over, a \textit{slot type} model is executed to predict type of the
answer from a set of \{\texttt{Yes}, \texttt{No}, \texttt{DontCare},
\texttt{Span}\}. If the \textit{slot type} model predicts \texttt{span},
\textit{slot span} model will finally be predicting the slot value as a span of
tokens within the dialog. The full model architecture is shown in
Figure~\ref{fig:architecture}.

\paragraph{Slot Carryover Model}
To model dynamic nature of dialog state, we introduce a model whose purpose is
to decide whether to carry over a slot value from the previous turn. For a given
slot $s_j$, $C_j(t) = 1$ if $s_j(t) \neq s_j(t-1)$ and $0$ if they are equal. We
multiply the dialog embedding $\textbf{e}(t)$ with a fully connected layer
$\textbf{W}_i$ to predict the change for slot $i$ as:
\begin{equation}
    P(C_i(t)) = sigmoid (\textbf{e}(t) \cdot \textbf{W}_i)
\end{equation} 

\noindent The network architecture is shown in Figure~\ref{fig:architecture}. In
our implementation, the weights $\textbf{W}_i$ for each slot are trained
together, i.e., the neural network would predict the slot carryover change
$C_i(t)$ jointly for all $M$ slots.

\paragraph{Slot Type Model}
A typical dialog state comprises of slots that can have both categorical and
named entities within the context of conversation. To adopt a flexible approach
and inspired by the state-of-the-art reading comprehension approaches, we
propose a classifier that predicts the type of slot value at each turn. In our
setting, we prescribe the output space to be \{\texttt{Yes}, \texttt{No},
\texttt{DontCare}, \texttt{Span}\} where \texttt{Span} indicates the slot value
is a named entity which can be found within the dialog. As shown in
Figure~\ref{fig:architecture}, we concatenate the dialog embedding
$\textbf{e}(t)$ with the question encoding $\textbf{q}_i$ for slot $i$ as the
input to the affine layer $\textbf{A}$ to predict the slot type $T_i(t)$ as:
\begin{equation}
    P(T_i(t)) \propto \exp(\textbf{A} \cdot (\textbf{e}(t);\textbf{q}_i))
\end{equation}

	%It is worth-noting that we predict the slot types independently for each question $\textbf{q}_i$ in contrast to the slot carryover model, where we predict the carryover for all slots at once.

\paragraph{Slot Span Model}
We map our slot values into a span with start and end position in our flattened
conversation $D_t$. We then use the dialog encoding vectors $\{\textbf{d}_1,
\textbf{d}_2, \dots \textbf{d}_{L}\}$ and the question vector $\textbf{q}_i$ to
compute the bilinear product and train two classifiers to predict the start
position and end position of the slot value. More specifically, for slot $j$, 
\begin{equation} 
    P_{j}^{(start)}(x) = \frac{\exp{(\textbf{d}_{x}  \Theta^{(start)} \textbf{q}_j)}}{\sum_{x'} \exp{(\textbf{d}_{x'}  \Theta^{(start)} \textbf{q}_j)}}
\end{equation}

\noindent Similarly, we define $P_{j}^{(end)}(x)$ with $\Theta^{(end)}$. During
span inference, we choose the best span from word $i$ to word ${i'}$ such that
$i \le i'$ and $P_j^{(start)}(i) \times P_j^{(end)}(i')$ is maximized, in line
with the approach by \citet{chen2017reading}.

%Describe the three components here: State Carryover(SC), Slot Type (ST) and Slot Span (SS). Mention that the context can be shared between the three models and we can undertake multi-task training. Also include a figure for the model architecture here (Ask Tag for the tool)

\section{Experiments} 
\subsection{Data}  
%Discuss the properties of the multiwoz dataset. Why did we choose it? What are the challenges for this dataset? mention the oracle accuracy etc. Indicate the annotation errors and give a hard example(discuss with Dilek)
We use the recently-released MultiWOZ-2.0 dataset
\cite{budzianowski2018multiwoz, ramadan2018large} to test our approach. This
dataset consists of multi-domain conversations from seven domains with a total
of 37 slots across domains. Many of these slot types such as \texttt{day} and
\texttt{people} are shared across multiple domains. In our experiments, we
process each slot independently by considering the concatenation of slot domain,
slot category, and slot name, e.g.,
\{\texttt{bus}.\texttt{book}.\texttt{people}\},
\{\texttt{restaurant}.\texttt{semi}.\texttt{food}\}. An example of conversation
is shown in Table~\ref{table:conv}. We use standard training/development/test
present in the data set. 

It is worth-noting that the dataset in the current form has certain annotation
errors. First, there is lack of consistency between the slot values in the
ontology and the ground truth in the context of the dialog. For example, the
ontology has \textit{moderate} but the dialog context has \textit{moderately}.
Second, there are erroneous delay in the state updates, sometimes extending
turns in the dialog. This error negatively impacts the performance
of the slot carryover model.

\subsection{Experimental Setup}
We train our three models independently without sharing the dialog context. For
all the three models, we encode the word tokens with BERT~\cite{devlin2018bert}
followed by an affine layer with 200 hidden units. This output is then fed into
a one-layer bi-directional LSTM with 50 hidden units to obtain the contextual
representation as show in Figure~\ref{fig:architecture}. In all our experiments,
we keep the parameters of the BERT embeddings frozen.

For slot carryover model, we predict a binary vector over 37 slots jointly to
get the decisions of whether to carry over values for each slot. For slot type
and slot span models, we treat dialog--question pairs ($D_t$, $q_i$) as separate
prediction tasks for each slot.

We use the learning rate of 0.001 with ADAM optimizer and batch size equal to 32
for all three models. We stop training our models when the loss on the
development set has not been decreasing for ten epochs.

\section{Results}
\begin{table}[]
	\centering
	\begin{tabular}{l|l}    
		Method         &  Accuracy      \\ \hline
		MultiWOZ Benchmark & 25.83\% \\
		GLAD \cite{zhong2018global}  & 35.57\% \\
		GCE \cite{nouri2018toward}  & 35.58\%   \\ 
		Our approach (single) & 39.41\%  \\ 
		Our approach (ensemble)  & 42.12\% \\
        HyST (ensemble) \cite{anonymous} & 44.22\% \\
		\textbf{Our approach + JST (ensemble)} & \textbf{47.33}\% \\
	\end{tabular}$  $
	\caption{Joint goal accuracy on MultiWOZ-2.0. We present both single and ensemble results for our approach.} \label{table:overall}
\end{table}
 %Abhishek: Dilek wants us to define the metric
Table~\ref{table:overall} presents our results on MultiWOZ-2.0 test dataset. We
compare our methods with global-local self-attention model
(GLAD)~\cite{zhong2018global}, global-conditioned encoder model
(GCE)~\cite{nouri2018toward}, and hybrid joint state tracking model
(OV ST+JST)~\cite{liu2017end, anonymous}. As in previous work, we report joint goal
accuracy as our metric. For each user turn, joint goal accuracy checks whether
all predicted states exactly matches the ground truth state for all slots.  We
can see that our system with single model can achieve 39.41\% joint goal
accuracy, and with the ensemble model we can achieve 42.12\% joint goal
accuracy.

%Abhishek: Dilek commented that this is too subjective, I will remove it. 
%In addition, it is worth noting that our model is much
%simpler and less complex than other methods presented in
%Table~\ref{table:overall}. 

Table~\ref{table:slot_acc} shows the accuracy for each slot type for both our
method and the joint state tracking approach with fix vocabulary in ~\citet{anonymous}. 
We can see our approach tends to have higher accuracy on some of
the slots that have larger set of possible values
such as \texttt{attraction.semi.name} and
\texttt{taxi.semi.destination}. 
%Abhishek: Dilek wants us to not write JST (as it is subjective)
However, it is worth-noting that even for slots with smaller vocabulary sizes
such as \texttt{hotel.book.day} and \texttt{hotel.semi.pricerange}, our approach
achieves better accuracy than using closed vocabulary approach. Our hypothesis
for difference is that such information appear more frequently in user utterance
thus our model is able to learn it more easily from the dialog context. 
%Whereas for
%slot \texttt{restaurant.semi.name}, JST has the higher accuracy due to the lack
%of entity resolution in our methods.

We also reported the result for a hybrid model by combining our approach with
the JST approach in ~\cite{anonymous}. Our combination strategy is as follows: first
we calculated the slot type accuracy for each model on
the development dataset; then for each slot type, we choose to use the
predictions from either our model or JST model based on the accuracy calculated
on the development set, whichever is higher. 
With this approach, we achieve the joint-goal accuracy of 46.28\%. We
hypothesize that this is because our method uses an open vocabulary, where all
the possible values can only be obtained from the conversation; the joint state
tracking method uses closed ontology, we can get the best of both the worlds by
combining two methods. 

\begin{table}[t!]
\setlength{\tabcolsep}{6pt} % Default value: 6pt
	\begin{center}
	\scalebox{0.82}{
		\begin{tabular}{l|llr}    
				\hline \bf Slot Name & \bf Ours  & \bf JST  & \textbf{Vocab} \\ 
				& & & \textbf {Size} \\ \hline
			attraction.semi.area &0.9637 &0.9719 &16 \\  
			\textbf{attraction.semi.name}& 0.9213 & 0.9013 &137 \\
			attraction.semi.type  & 0.9205 & 0.9637 &37\\
			bus.book.people  & 1.0000 & 1.0000 &1\\
			bus.semi.arriveBy & 1.0000 & 1.0000  &1\\
			bus.semi.day  & 1.0000 & 1.0000 &2\\
			bus.semi.departure & 1.0000 & 1.0000 &2 \\
			bus.semi.destination  & 1.0000 & 1.0000 &5\\
			bus.semi.leaveAt & 1.0000 & 1.0000 &2 \\
			\bf hospital.semi.department & 0.9991 & 0.9988 &52 \\
			\bf hotel.book.day  & 0.9863 & 0.9784 &11\\
			hotel.book.people & 0.9714 & 0.9847 &9 \\
			hotel.book.stay & 0.9736 & 0.9809 &9 \\
			\bf hotel.semi.area & 0.9679 & 0.9570 &24\\
			hotel.semi.internet & 0.9713 & 0.9718 &8\\
			\bf hotel.semi.name & 0.9147 & 0.9056 &89\\
			hotel.semi.parking & 0.9563 & 0.9657 &8\\
			\bf hotel.semi.pricerange & 0.9679 & 0.9666 &9\\
			hotel.semi.stars  & 0.9627 & 0.9759 &13\\
			hotel.semi.type & 0.9140 & 0.9261 &18\\
			\bf restaurant.book.day & 0.9874 & 0.9871  &10\\
			restaurant.book.people  & 0.9787 & 0.9881 &9\\
			\bf restaurant.book.time  & 0.9882 & 0.9578 &61\\
			restaurant.semi.area  & 0.9607 & 0.9654 &19\\
			\bf restaurant.semi.food & 0.9741 & 0.9691 &104\\
			\bf restaurant.semi.name & 0.9113 & 0.8781 &183\\
			\bf restaurant.semi.pricerange & 0.9662 & 0.9626 &11\\
			\bf taxi.semi.arriveBy & 0.9893 & 0.9719 &101\\
			\bf taxi.semi.departure & 0.9665 & 0.9304 &261\\
			\bf taxi.semi.destination & 0.9634 & 0.9288 &277\\
			\bf taxi.semi.leaveAt & 0.9821 & 0.9524 &119\\
			train.book.people & 0.9586 & 0.9718 &13\\
			\bf train.semi.arriveBy & 0.9738 & 0.9491 &107\\
			\bf train.semi.day & 0.9854 & 0.9783 &11\\
			train.semi.departure & 0.9599 & 0.9710 &35\\
			train.semi.destination & 0.9538 & 0.9699 &29\\
			\bf train.semi.leaveAt & 0.9595 & 0.9478 &134\\
			\hline
		\end{tabular}}
	\end{center}
	\caption{\label{table:slot_acc} Slot accuracy breakdown for our approach
versus joint state tracking method. Bolded slots are the ones have better performance using our attentive reading comprehension approach. }
\end{table}

\subsection{Ablation Analysis}

\begin{table}[]
	\centering
	\begin{tabular}{l|lll}    
		 Ablation    &  Dev  Accuracy      \\ \hline
 		Oracle Models & 73.12\%\\ %71.93\% \\ 
		Our approach & 41.10\% \\ 
		- BERT & 39.19\% \\ 
		+ Oracle Slot Type Model  & 41.43\%   \\  
		+ Oracle Slot Span Model &  45.77\% \\  
		+ Oracle Slot Carryover Model & 60.18\%  \\ 
	\end{tabular}
	\caption{Ablation study on our model components for MultiWOZ-2.0 on development set for joint goal accuracy.}\label{table:ablation}
\end{table}
%Tag: slightly better description of oracle model is needed
Table~\ref{table:ablation} illustrates the ablation studies for our model on
development set. The contextual embedding BERT~\cite{devlin2018bert} can give us
around 2\% gains. As for the oracle models, we can see that even if using all
the oracle results (ground truth), our development set accuracy is only 73.12\%. This is because our approach is only
considering the values within the conversation, if values are not present in the
dialog, the oracle models would fail. It is interesting to see that if we
replace our slot carryover model with an oracle one, the accuracy improves significantly
to 60.18\% (+19.08\%) compared to replacing other two models (41.43\% and
45.77\%). This is because our span-based reading comprehension approach model already gives us accuracy as high as 96\% per slot on development data, there is not much room for improvement. Whereas our binary slot carryover model only achieve an accuracy of 72\% per turn. We hypothesis that for slot carryover problem is imbalanced, i.e., there are significantly more slot carryovers than slot updates, making the model training and predictions harder. This suggest further improvements are needed for \textit{slot carryover} model to make overall state tracking accuracy higher. 

\subsection{Error Analysis}
In Table~\ref{table:slotspan}, we conduct an error analysis of our models and
investigate its performance for different use cases. Since we formulate the
problem to be an open-vocabulary state tracking approach wherein the slot values are extracted in the dialog context, we divide the errors into following categories:
\begin{table*}[t]
	\centering
	\begin{tabular}{p{2.4cm}| p{1.9cm} |p{1.9cm} |p{7cm} |p{1cm}} 
		Category     &  Hypothesis & Reference & Context & (\%)   \\ \hline
%\multirow{2}{*}{Unanswerable Slot} & NONE & not NONE \\ \cline{1-2} not NONE & NONE \\ \hline
\multirow{2}{2cm}{Unanswerable Slot Error} & not \textit{None} & \textit{None}& \dots & 42.4 \\\cline{2-5} & \textit{None} & not \textit{None} & \dots & 23.1 \\ \hline 

		Incorrect slot Reference &4&8& \dots 3 nights, and \textbf{4 people}.  Thank You! 
		[A] Booking was unsuccessful \dots I'd like to book there Monday for 1 night with \textbf{8 people}.  \dots & 19.1 \\ \hline
		Incorrect Slot Resolution & 3:30 & 15:30 & \dots you like to arrive at the Cinema? [U] I want to leave the hotel by \textbf{3:30} [A] Your taxi reservation departing \dots & 12.9 \\ \hline 
		Imprecise Slot Boundary  & nandos city centre  & nandos &  \dots number is 01223902168 [U] Great I am also looking for a restaurant called \textbf{nandos city centre} \dots & 2.5 \\ \hline  

	\end{tabular}
	\caption{Error categorization and percentage distribution: representative example from each category and an estimate breakdown of the error types on development set, based on the analysis of 200 error samples produced by our model. Numbers of the first category is exact because we are able to summarize this error category statistically.}\label{table:slotspan}
	%	\vspace{2mm}
\end{table*}

\begin{itemize}
	\item \textbf{Unanswerable Slot Error} This category contains two type of errors: (1) Ground truth slot is a not \textit{None} value, but our prediction is \textit{None}; (2) Ground truth slot is \textit{None}, but our prediction is a not \textit{None} value. This type of error can be attributed to the incorrect predictions made by our slot carryover model.

	\item \textbf{Imprecise Slot Reference} where multiple potential candidates in the context exists. The model refers to the incorrect \textit{entity} in the conversation. This error can be largely attributed to following reasons: (1) the model overfits to the set of tokens that it has seen more frequently in the training set; (2) the model does not generalize well for scenarios where the user corrects the previous entity; (3) the model incorrectly overfits to the order or position of the entity in the context. These reasons motivate future research in incorporating more neural reading comprehension approaches for dialog state tracking.  
	\item \textbf{Imprecisie Slot Resolution}  In this type of errors, we cannot find the exact match of ground truth value in the dialog context. However, our predicted model span is a paraphrase or has very close meaning to the ground truth. This error is inherent in approaches that do not
	extract the slot value from an ontology but rather the dialog context. On
	similar lines, we also observe cases where the slot value in the dialog context
	is \textit{resolved} (or canonicalized) to a different surface-form entity that
	is perhaps more amenable for downstream applications.

	\item \textbf{Imprecise Slot Boundary} In this category of errors, our model
	chooses a span that is either a superset or subset of the correct reference.
	This error is especially frequent for proper nouns where the model has a weaker
	signal to outline the slot boundary precisely. 

%It is worth noting that
%the model correctly generalizes to correct values even with infrequent
%occurrences for these slot values in the training data.
 %in which the predicted slot value is a paraphrase of the ground truth slot value. 

%	\item \textbf{Data Annotation} errors where either the slot reference or the value in dialog context is incorrect due to annotation noise. 
		
\end{itemize}

Table~\ref{table:slotspan} provides us the error examples and estimated percentage from each category. "Unanswerable Slot" accounts for 65.5\% errors for our model, this indicates further attention may be needed to the slot carryover model, otherwise it would become a barrier even if we have a perfect span model. This finding is in alignment with our ablation studies in Table~\ref{table:ablation}, where oracle slot carryover model would give us the most boost in joint goal accuracy. Additionally, 12.9\% of errors are due to imprecise slot resolution, this suggests future directions of resolving the context words to the ontology.

\subsection{Evaluating Different Context Encoders for Slot Carryover Model}  
% Tag: more detailed experiments with hierarchical context encoder.
As shown in oracle ablation studies in Table~\ref{table:ablation}, slot carryover model plays a significant role in our pipeline. Therefore we explore the different types of context encoders for slot carryover model to see whether if it improves the performance in table~\ref{table:features}.
 In addition to use a flat dialog context of user and agent turns
\texttt{[U]} and \texttt{[A]} to predict carryover for every slot in the state, 
we explored hierarchical context encoder with an utterance-level LSTM over each user and agent utterance and a dialog-level LSTM over the whole dialog with both constrained and unconstrained context window, similar to~\citet{liu2017end}. However, we did not witness any significant
performance change across the two variants as show in Table~\ref{table:features}. Lastly, we employed self-attention
over the flattened dialog context in line with \citet{vaswani2017attention}. However, we can see from Table~\ref{table:features} that this strategy slightly hurts the model performance. One hypothesis for sub par slot carryover model performance is due to the inherent noise in the annotated data for state updates. Through a preliminary analysis on the development set, we encountered few erroneous delay in the state updates sometimes extending to over multiple turns. Nevertheless, these experimental results motivate future research in \textit{slot carryover} models for multi-domain conversations.

\begin{table}[]  
	\centering
	\begin{tabular}{p{5cm}|p{1.6cm}}    
		Context Feature & Per Turn Carryover Accuracy\\
		\hline 
	Flat Context (LSTM) & 75.10\% \\ 
	Hierarchical Context (all turns) & 75.98\% \\
	Hierarchical Context ($\le$ 3 turns) &75.60\% \\
	Flat Context (Self-Attention)  &  74.75\% \\
	\end{tabular}
	\caption{Analyzing the different types of context features for Slot Carryover Model}\label{table:features}
	%\vspace{2mm}
\end{table}

\subsection{Analyzing Conversation Depth} 
\begin{table}[]  
	\centering
	\begin{tabular}{c|c|c}    
		Conversation & Total & \% Incorrect  \\ 
		Depth $t$ & Turns &  Turns \\
		\hline 
1&1000&23.90\\
2&1000&38.30\\
3&997&50.85\\
4&959&61.52\\
5&892&71.52\\
6&811&76.82\\
7&656&82.77\\
8&475&87.37\\
9&280&89.64\\
10&153&94.77\\
	\end{tabular}
	\caption{Analyzing the overall model robustness for conversation depth for MultiWOZ-2.0}\label{table:depth}
	\vspace{2mm}
\end{table}

In Table~\ref{table:depth}, we explore the relationship between the depth of a conversation and the performance of our models. 
%Abhishek: Dilek wants us to write more about defining incorrect turns
More precisely, we segment a given set of dialogs into individual \textit{turns} and measure the state accuracy for each of these segments. We mark a turn correct only if all the slots in its state are predicted correctly. We observe that the model performance degrades as the number of turns increase. The primary reason for this behavior is that an error committed earlier in the
conversation can be carried over for later turns. This results in a strictly higher probability for a later turn to be incorrect as compared to the turns
earlier in the conversation. These results motivate future research in
formulating models for state tracking that are more robust to the depth of the conversation.

%\subsubsection{Single domain vs Multi Domain Context}
%\begin{table}[]  
%	\centering
%	\begin{tabular}{c|c|c}    
%		Number of & Total & Incorrect  \\ 
%		domains $d$ & Turns &  \ Turns $\hat{\epsilon}(t)$\\
%		\hline 
%		0 & & \\
%		1 & &\\  
%		2 &  &\\ 
%		2+ & &\\ 
%	\end{tabular}
%	\caption{Analyzing model robustness for conversation depth for MultiWOZ-2.0}\label{table:depth}
%	\vspace{2mm}
%\end{table}

%\begin{table}[]\label{answers}
%	\centering
%	\begin{tabular}{l|l|l}    
%		Answer Type   & freq & Acc \\ \hline
%		Yes &  &  \\ \
%		No &  &  \\ 
%		Dont Care & &   \\ 
%		Named Entity (I) & & 92\% 
%	\end{tabular} 
%	\caption{Model performance on answer types for MultiWOZ-2.0}
%	\vspace{2mm}
%\end{table}

\section{Conclusion}

The problem of tracking user's belief state in a dialog is a historically significant endeavor. In that context, research on dialog state tracking has been geared towards discriminative methods, where these methods are usually estimating the distribution of user state over a fixed vocabulary. However, modern dialog systems presents us with problems requiring a large scale perspective. It is not unusual to have thousands of slot values in the vocabulary which could have millions variations of dialogs. So we need a vocabulary-free way to pick out the slot values. 

How can we pick the slot values given an infinite amount of vocabulary size? Some methods adopt a candidate generation mechanism to generate slot values and make a binary decision with the dialog context. Attention-based neural network gives a clear and general basis for selecting the slot values by direct pointing to the context spans. While this type of methods has already been proposed recently, we explored this type of idea furthermore on MultiWOZ-2.0 dataset.

We introduced a simple attention based neural network to encode the dialog context and point to the slot values within the conversation. We have also introduced an additional slot carryover model and showed its impact on the model performance. By incorporating the deep contextual word embeddings and combining the traditional fixed vocabulary approach, we significantly improved the joint goal accuracy on MultiWOZ-2.0.
 
 We also did a comprehensive analysis to see to what extent our proposed model can achieve. One interesting and significant finding from the oblation studies suggests the importance of the slot carryover model. We hope this finding can inspire future dialog state tracking research to work towards this direction, i.e., predicting whether a slot of state is none or not.

The field of machine reading comprehension has made significant progress in recent years. We believe human conversation can be viewed as a special type of context and we hope that the developments suggested here can help dialog related tasks benefit from modern reading comprehension models.

%The problem of machine language understanding is a historically significant endeavor. One of its task, machine reading comprehension, has been made significant progress

%Recent advances in neural networks have shown impressive progress on machine reading comprehension task. 

%In this paper, we discussed a novel approach to inferring the dialog state by
%treating prior context in the conversation to reading a passage. We employ an
%extension of the reading comprehension approach and augment it with the state
%carryover and slot type models to address the inherent dynamic flow of
%information in a dialog system. Since our model is not limited by the candidate
%values for a slot, it is scalable for both unbounded value sets and unseen slot
%values. To test the approach, we used the MultiWOZ-2.0 dataset and obtained a
%joint-goal accuracy of 46.28\% with the closed vocabulary approach JST, exceeding current state-of-the-art by 10.7\%\@.
%In addition, we analyzed the slot-level performance, the relative importance of
%each prediction model, the robustness of the model with regards to the depth of the
%conversation. Finally we categorized the common errors incurred by our slot span
%model showed what our model does well and where it tends fai

\section*{Acknowledgements}
The authors would like to thank Rahul Goel, Shachi Paul, Anuj Kumar Goyal, Angeliki Metallinou and other Alexa AI team members for their useful discussions and feedbacks.

\newpage
\bibliography{all}

\begin{thebibliography}{28}
\expandafter\ifx\csname natexlab\endcsname\relax\def\natexlab#1{#1}\fi

\bibitem[{Budzianowski et~al.(2018)Budzianowski, Wen, Tseng, Casanueva, Stefan,
  Osman, and Ga{\v{s}}i\'c}]{budzianowski2018multiwoz}
Pawe{\l} Budzianowski, Tsung-Hsien Wen, Bo-Hsiang Tseng, I{\~n}igo Casanueva,
  Ultes Stefan, Ramadan Osman, and Milica Ga{\v{s}}i\'c. 2018.
\newblock Multiwoz - a large-scale multi-domain wizard-of-oz dataset for
  task-oriented dialogue modelling.
\newblock In \emph{Proceedings of the 2018 Conference on Empirical Methods in
  Natural Language Processing (EMNLP)}.

\bibitem[{Chen(2018)}]{chen2018neural}
Danqi Chen. 2018.
\newblock \emph{Neural Reading Comprehension and Beyond}.
\newblock Ph.D. thesis, Stanford University.

\bibitem[{Chen et~al.(2017)Chen, Fisch, Weston, and Bordes}]{chen2017reading}
Danqi Chen, Adam Fisch, Jason Weston, and Antoine Bordes. 2017.
\newblock Reading {Wikipedia} to answer open-domain questions.
\newblock In \emph{Association for Computational Linguistics (ACL)}.

\bibitem[{Devlin et~al.(2018)Devlin, Chang, Lee, and
  Toutanova}]{devlin2018bert}
Jacob Devlin, Ming-Wei Chang, Kenton Lee, and Kristina Toutanova. 2018.
\newblock Bert: Pre-training of deep bidirectional transformers for language
  understanding.
\newblock \emph{arXiv preprint arXiv:1810.04805}.

\bibitem[{Goel et~al.(2018)Goel, Paul, Chung, Lecomte, Mandal, and
  Hakkani-Tur}]{goel2018flexible}
Rahul Goel, Shachi Paul, Tagyoung Chung, Jeremie Lecomte, Arindam Mandal, and
  Dilek Hakkani-Tur. 2018.
\newblock Flexible and scalable state tracking framework for goal-oriented
  dialogue systems.
\newblock \emph{arXiv preprint arXiv:1811.12891}.

\bibitem[{Goel et~al.(2019)Goel, Paul, and Hakkani-T{\"u}r}]{anonymous}
Rahul Goel, Shachi Paul, and Dilek Hakkani-T{\"u}r. 2019.
\newblock Hyst: A hybrid approach for flexible and accurate dialogue state
  tracking.

\bibitem[{Hermann et~al.(2015)Hermann, Kocisky, Grefenstette, Espeholt, Kay,
  Suleyman, and Blunsom}]{hermann2015teaching}
Karl~Moritz Hermann, Tomas Kocisky, Edward Grefenstette, Lasse Espeholt, Will
  Kay, Mustafa Suleyman, and Phil Blunsom. 2015.
\newblock Teaching machines to read and comprehend.
\newblock In \emph{Advances in neural information processing systems}, pages
  1693--1701.

\bibitem[{Ko{\v{c}}isk{\`y} et~al.(2018)Ko{\v{c}}isk{\`y}, Schwarz, Blunsom,
  Dyer, Hermann, Melis, and Grefenstette}]{kovcisky2018narrativeqa}
Tom{\'a}{\v{s}} Ko{\v{c}}isk{\`y}, Jonathan Schwarz, Phil Blunsom, Chris Dyer,
  Karl~Moritz Hermann, G{\'a}abor Melis, and Edward Grefenstette. 2018.
\newblock The narrativeqa reading comprehension challenge.
\newblock \emph{Transactions of the Association of Computational Linguistics},
  6:317--328.

\bibitem[{Lai et~al.(2017)Lai, Xie, Liu, Yang, and Hovy}]{lai2017race}
Guokun Lai, Qizhe Xie, Hanxiao Liu, Yiming Yang, and Eduard Hovy. 2017.
\newblock Race: Large-scale reading comprehension dataset from examinations.
\newblock \emph{arXiv preprint arXiv:1704.04683}.

\bibitem[{Liu and Lane(2017)}]{liu2017end}
Bing Liu and Ian Lane. 2017.
\newblock An end-to-end trainable neural network model with belief tracking for
  task-oriented dialog.
\newblock \emph{Proc. Interspeech 2017}, pages 2506--2510.

\bibitem[{Liu et~al.(2019)Liu, Gardner, Belinkov, Peters, and
  Smith}]{liu2019linguistic}
Nelson~F Liu, Matt Gardner, Yonatan Belinkov, Matthew Peters, and Noah~A Smith.
  2019.
\newblock Linguistic knowledge and transferability of contextual
  representations.
\newblock \emph{arXiv preprint arXiv:1903.08855}.

\bibitem[{Nouri and Hosseini-Asl(2018)}]{nouri2018toward}
Elnaz Nouri and Ehsan Hosseini-Asl. 2018.
\newblock Toward scalable neural dialogue state tracking model.
\newblock In \emph{32nd Conference on Neural Information Processing Systems
  (NeurIPS 2018), 2nd Conversational AI workshop}.

\bibitem[{Perez and Liu(2017)}]{perez2017dialog}
Julien Perez and Fei Liu. 2017.
\newblock Dialog state tracking, a machine reading approach using memory
  network.
\newblock In \emph{Proceedings of the 15th Conference of the European Chapter
  of the Association for Computational Linguistics: Volume 1, Long Papers},
  pages 305--314.

\bibitem[{Peters et~al.(2018)Peters, Neumann, Iyyer, Gardner, Clark, Lee, and
  Zettlemoyer}]{peters2018deep}
Matthew~E Peters, Mark Neumann, Mohit Iyyer, Matt Gardner, Christopher Clark,
  Kenton Lee, and Luke Zettlemoyer. 2018.
\newblock Deep contextualized word representations.
\newblock \emph{arXiv preprint arXiv:1802.05365}.

\bibitem[{Rajpurkar et~al.(2016)Rajpurkar, Zhang, Lopyrev, and
  Liang}]{rajpurkar2016squad}
Pranav Rajpurkar, Jian Zhang, Konstantin Lopyrev, and Percy Liang. 2016.
\newblock Squad: 100,000+ questions for machine comprehension of text.
\newblock In \emph{Proceedings of the 2016 Conference on Empirical Methods in
  Natural Language Processing}, pages 2383--2392.

\bibitem[{Ramadan et~al.(2018)Ramadan, Budzianowski, and
  Gasic}]{ramadan2018large}
Osman Ramadan, Pawe{\l} Budzianowski, and Milica Gasic. 2018.
\newblock Large-scale multi-domain belief tracking with knowledge sharing.
\newblock In \emph{Proceedings of the 56th Annual Meeting of the Association
  for Computational Linguistics}, volume~2, pages 432--437.

\bibitem[{Rastogi et~al.(2017)Rastogi, Hakkani-T{\"u}r, and
  Heck}]{rastogi2017scalable}
Abhinav Rastogi, Dilek Hakkani-T{\"u}r, and Larry Heck. 2017.
\newblock Scalable multi-domain dialogue state tracking.
\newblock In \emph{Automatic Speech Recognition and Understanding Workshop
  (ASRU), 2017 IEEE}, pages 561--568. IEEE.

\bibitem[{Reddy et~al.(2019)Reddy, Chen, and Manning}]{reddy2019coqa}
Siva Reddy, Danqi Chen, and Christopher~D Manning. 2019.
\newblock {CoQA}: A conversational question answering challenge.
\newblock \emph{Transactions of the Association of Computational Linguistics
  (TACL)}.

\bibitem[{Richardson et~al.(2013)Richardson, Burges, and
  Renshaw}]{richardson2013mctest}
Matthew Richardson, Christopher~JC Burges, and Erin Renshaw. 2013.
\newblock Mctest: A challenge dataset for the open-domain machine comprehension
  of text.
\newblock In \emph{Proceedings of the 2013 Conference on Empirical Methods in
  Natural Language Processing}, pages 193--203.

\bibitem[{Seo et~al.(2016)Seo, Kembhavi, Farhadi, and
  Hajishirzi}]{seo2016bidirectional}
Minjoon Seo, Aniruddha Kembhavi, Ali Farhadi, and Hannaneh Hajishirzi. 2016.
\newblock Bidirectional attention flow for machine comprehension.
\newblock \emph{arXiv preprint arXiv:1611.01603}.

\bibitem[{Sutskever et~al.(2014)Sutskever, Vinyals, and
  Le}]{sutskever2014sequence}
Ilya Sutskever, Oriol Vinyals, and Quoc~V Le. 2014.
\newblock Sequence to sequence learning with neural networks.
\newblock In \emph{Advances in neural information processing systems}, pages
  3104--3112.

\bibitem[{Vaswani et~al.(2017)Vaswani, Shazeer, Parmar, Uszkoreit, Jones,
  Gomez, Kaiser, and Polosukhin}]{vaswani2017attention}
Ashish Vaswani, Noam Shazeer, Niki Parmar, Jakob Uszkoreit, Llion Jones,
  Aidan~N Gomez, {\L}ukasz Kaiser, and Illia Polosukhin. 2017.
\newblock Attention is all you need.
\newblock In \emph{Advances in neural information processing systems}, pages
  5998--6008.

\bibitem[{Williams et~al.(2005)Williams, Poupart, and
  Young}]{williams2005factored}
Jason~D Williams, Pascal Poupart, and Steve Young. 2005.
\newblock Factored partially observable markov decision processes for dialogue
  management.
\newblock In \emph{Proc. IJCAI Workshop on Knowledge and Reasoning in Practical
  Dialogue Systems}, pages 76--82.

\bibitem[{Williams and Young(2007)}]{williams2007partially}
Jason~D Williams and Steve Young. 2007.
\newblock Partially observable markov decision processes for spoken dialog
  systems.
\newblock \emph{Computer Speech \& Language}, 21(2):393--422.

\bibitem[{Wu et~al.(2019)Wu, Madotto, Hosseini-Asl, Xiong, Socher, and
  Fung}]{wu2019transferable}
Chien-Sheng Wu, Andrea Madotto, Ehsan Hosseini-Asl, Caiming Xiong, Richard
  Socher, and Pascale Fung. 2019.
\newblock Transferable multi-domain state generator for task-oriented dialogue
  systems.
\newblock In \emph{Proceedings of the 57th Annual Meeting of the Association
  for Computational Linguistics}, pages 808--819.

\bibitem[{Xu and Hu(2018)}]{xu2018end}
Puyang Xu and Qi~Hu. 2018.
\newblock An end-to-end approach for handling unknown slot values in dialogue
  state tracking.
\newblock In \emph{Proceedings of the 56th Annual Meeting of the Association
  for Computational Linguistics (ACL)}.

\bibitem[{Zhong et~al.(2018)Zhong, Xiong, and Socher}]{zhong2018global}
Victor Zhong, Caiming Xiong, and Richard Socher. 2018.
\newblock Global-locally self-attentive dialogue state tracker.
\newblock \emph{arXiv preprint arXiv:1805.09655}.

\bibitem[{Zhu et~al.(2018)Zhu, Zeng, and Huang}]{zhu2018sdnet}
Chenguang Zhu, Michael Zeng, and Xuedong Huang. 2018.
\newblock Sdnet: Contextualized attention-based deep network for conversational
  question answering.
\newblock \emph{arXiv preprint arXiv:1812.03593}.

\end{thebibliography}
\bibliographystyle{acl_natbib.bst}

\end{document}